\documentclass[a4paper, 10pt, conference]{ieeeconf}      

\IEEEoverridecommandlockouts                              

\usepackage{hyperref}

\usepackage{listings}

\usepackage[listings,skins]{tcolorbox}
\tcbuselibrary{breakable}

\newtcblisting{cpp}{
    listing engine=listings,
    breakable,
    listing options={
        language = C++,
        breaklines=true,
        columns=flexible, 
        keepspaces=true, 
        basicstyle=\footnotesize\ttfamily,
        keywordstyle=\color{blue},
        commentstyle=\itshape\color{violet},
        showstringspaces=false
    },
    listing only,
    colback=black!5,
    left=1mm,
    top=-1mm,
    bottom=-1mm,
    frame hidden,
    enhanced,
    borderline={0.5pt}{1pt}{black}
}

\title{\LARGE \bf
VERNIER: an open-source software pushing marker pose estimation down to the micrometer and nanometer scales
}

\author{Patrick Sandoz$^{1}$, Antoine N. André$^{2}$ and Guillaume J. Laurent$^{1}$ 
\thanks{$^{1}$Patrick Sandoz and Guillaume J. Laurent are with Université Marie et Louis Pasteur, SupMicroTech, CNRS, Institut FEMTO-ST, F-25000 Besançon, France
        {\tt\small patrick.sandoz@femto-st.fr, guillaume.laurent@supmicrotech.fr}}%
\thanks{$^{2}$Antoine N. André is with the National Institute of Advanced Industrial Science and Technology, CNRS-AIST JRL, IRL, Tsukuba, Japan
        {\tt\small antoine.andre@aist.go.jp}}%
\thanks{All authors contributed equally as first authors to this work.}
\thanks{This work has been supported by the ANR project Holo-Control (ANR-21-CE42-0009) and by the Bourgogne-Franche-Comté Region (Nano6D). It has been achieved in the frame of the EIPHI Graduate School (ANR-17-EURE-0002). The encoded targets were realized thanks to the RENATECH technological network and its FEMTO-ST facility MIMENTO. The experiments were conducted within the ROBOTEX network (ANR-21-ESRE-0015) and its FEMTO-ST technological facility CMNR.}
}

\begin{document}

\maketitle
\thispagestyle{empty}

\begin{abstract}
Pose estimation is still a challenge at the small scales. Few solutions exist to capture the 6 degrees of freedom of an object with nanometric and microradians resolutions over relatively large ranges. Over the years, we have proposed several fiducial marker and pattern designs to achieve reliable performance for various microscopy applications. Centimeter ranges are possible using pattern encoding methods, while nanometer resolutions can be achieved using phase processing of the periodic frames. This paper presents VERNIER, an open source phase processing software designed to provide fast and reliable pose measurement based on pseudo-periodic patterns. Thanks to a phase-based local thresholding algorithm, the software has proven to be particularly robust to noise, defocus and occlusion. The successive steps of the phase processing are presented, as well as the different types of patterns that address different application needs. The implementation procedure is illustrated with synthetic and experimental images. Finally, guidelines are given for selecting the appropriate pattern design and microscope magnification lenses as a function of the desired performance.
\end{abstract}



\section{Introduction}

Microrobotics aims to develop small robots, precision manipulators and automated machines able to handle, assemble and characterize micro and nano objects \cite{abbott2007robotics}. These tasks require to monitor the position and the orientation of end-effectors as in usual industrial robotics. 

At human scale, computer vision is widely used to track the movement of people and objects in many applications. Motion capture systems have become indispensable in film production, video game development, drone design and zoology \cite{nagy2023smart}. A lot of fiducial markers have been proposed to track or localize robots in various domains such as construction, logistic, space, and surgery \cite{kalaitzakis2021fiducial}. 

When the size of the object of interest decreases, pose estimation becomes challenging due to the lack of space to integrate sensors and due to the constraint of microscopy imaging. Unlike regular cameras, microscopes suffer from narrow fields of view (FoV), short depths of field, low contrasts and out-of-focus blurs, and usual fiducial markers performs poorly in these conditions.

There are few alternatives to imaging for measuring the six degrees of freedom (DoF) of a solid with nanometer and microradian resolutions. The most common option is to use six laser interferometers, each of which provides a measurement with sub-nanometer resolution in a single direction \cite{lee2011design, ortlepp2024high}. This setup is very cumbersome and requires complex calibration procedures. Furthermore, such approaches do not provide a sufficiently wide angular measurement range. Thus, computer vision remains an essential tool for estimating robot pose.

\begin{figure}
    \centering
    \includegraphics[width=\linewidth]{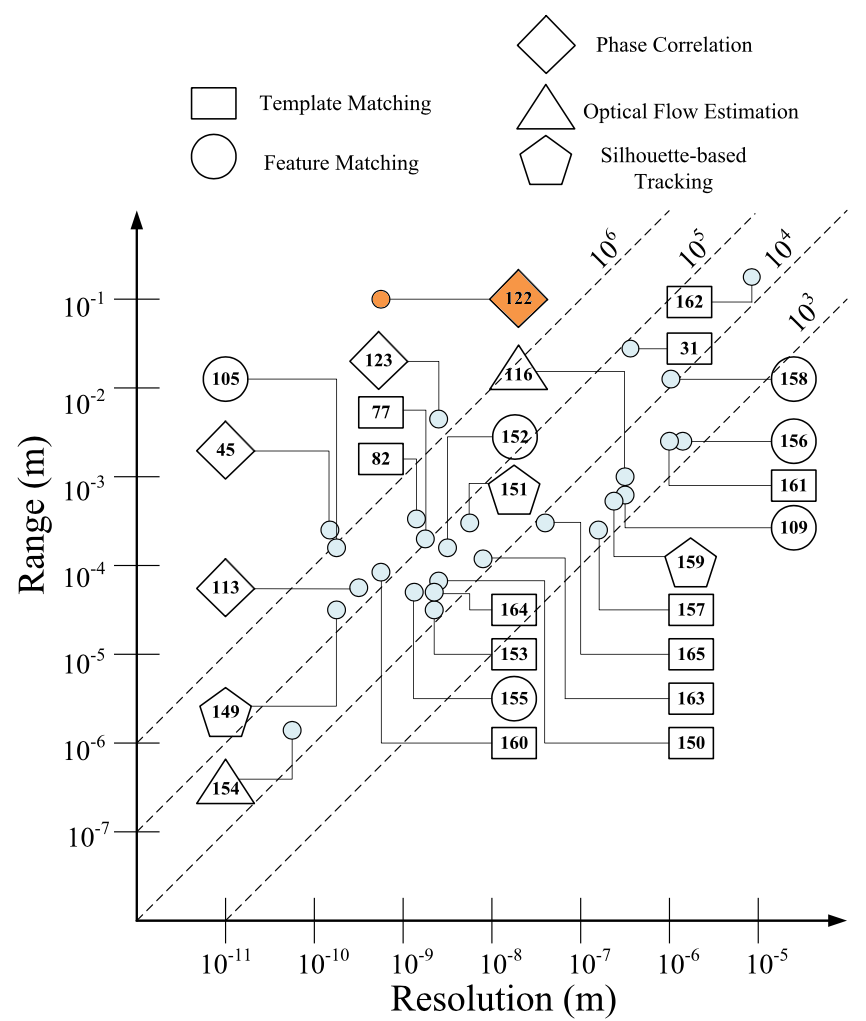}
    \caption{Comparison of vision-based methods dedicated to pose measurement under microscopy published in \cite{Yao2021review} (courtesy of IEEE). Numbers correspond to the references of this review. The method that is implemented in VERNIER is number 122 in orange at the top of the graph.}
    \label{fig.review}
\end{figure}

Many vision-based methods have been proposed to tackle pose estimation at the small scales. In 2021, the Fatikow's team published a review paper comparing the resolution and the range of state-of-the-art vision-based localization methods \cite{Yao2021review}. Fig.~\ref{fig.review} is extracted from this article and shows that most methods at these scales, rely on template matching. The most precise methods use phase correlation and can achieve sub-nanometer resolutions. However, their measurement ranges are still limited by the microscope's FoV. 

To overcome this limitation, pseudo-periodic patterns can be used to encode the absolute position over centimetric ranges, while using phase measurement to achieve nanometer resolutions \cite{andre2020sensing}. Based on this principle, we have designed several markers and patterns, along with dedicated processing algorithms, to ensure reliable performance in various microscopy applications. As it can be see in Fig.~\ref{fig.review}, this approach outperforms all others in terms of range-to-resolution ratio. Recently, we achieved the measurement of six DoF using these patterns with digital holographic microscopy that performs similarly to interferometer-based setups \cite{ahmad2024-6DoF}. 

\bigskip

This article presents the summary of ten years of research in the form of an open-source \texttt{C++} library, called VERNIER like the french mathematician Pierre Vernier who was the inventor and eponym of the vernier scale which increases the resolution of measuring devices like calipers. The VERNIER software provides a powerful tool to measure the 2D and 3D poses of objects observed through a microscope lens with unprecedented sub-pixel resolutions and ranges. The library works in combination with periodic markers fixed on the objects of interest, allowing accurate measurements even in noisy and blurred imaging contexts which commonly occurs in the microscopy context. The image processing involves two complementary steps, resulting on the one hand in fine but relative position measurement and, on the other hand, a coarser but absolute localization. The fine position process achieves high sub-pixel resolutions thanks to spectral analysis of the imaged periodic frame of the marker. Complementary coarse measurements are retrieved either from the marker contours or from a binary sequence encrypted within the periodic frame of the pattern.

VERNIER is based on different types of markers, designed to meet different application requirements, and whose design can be easily adapted to specific imaging parameters, mainly magnification, FoV, and expected measurement range and resolution. The next section introduces the different markers and summarizes the main image processing steps required to retrieve the associated position data. Section~\ref{sec:library} presents the software architecture and library functionality, while section~\ref{sec:applications} gives examples of applications using each marker type.
Finally, guidelines are proposed to enable future users to easily implement this measurement method and meet their specific needs. 
All sources and data are available on GitHub and can be downloaded for evaluation and testing\footnote{\url{https://github.com/vernierlib/vernier}}. 

\section{Marker types and measurement principle}\label{sec:markerTypes}

\subsection{Marker types}
VERNIER is designed for two families of markers: small and large markers. Small markers are intended to remain entirely within the FoV of the imaging system. Large markers are much wider than the FoV of the imaging system and allow to retrieve the absolute position of the observed area from the binary decoding of the local image captured, with respect to the entire pattern. Small markers are used for measuring short displacements as well as for differential measurements between several markers appearing in the same image. Large markers make the FoV and the measurement range independent of each other, allowing very large range-to-resolution ratios to be achieved, of up to $10^8$. 

\begin{figure}
    \centering
    \includegraphics[width = \linewidth]{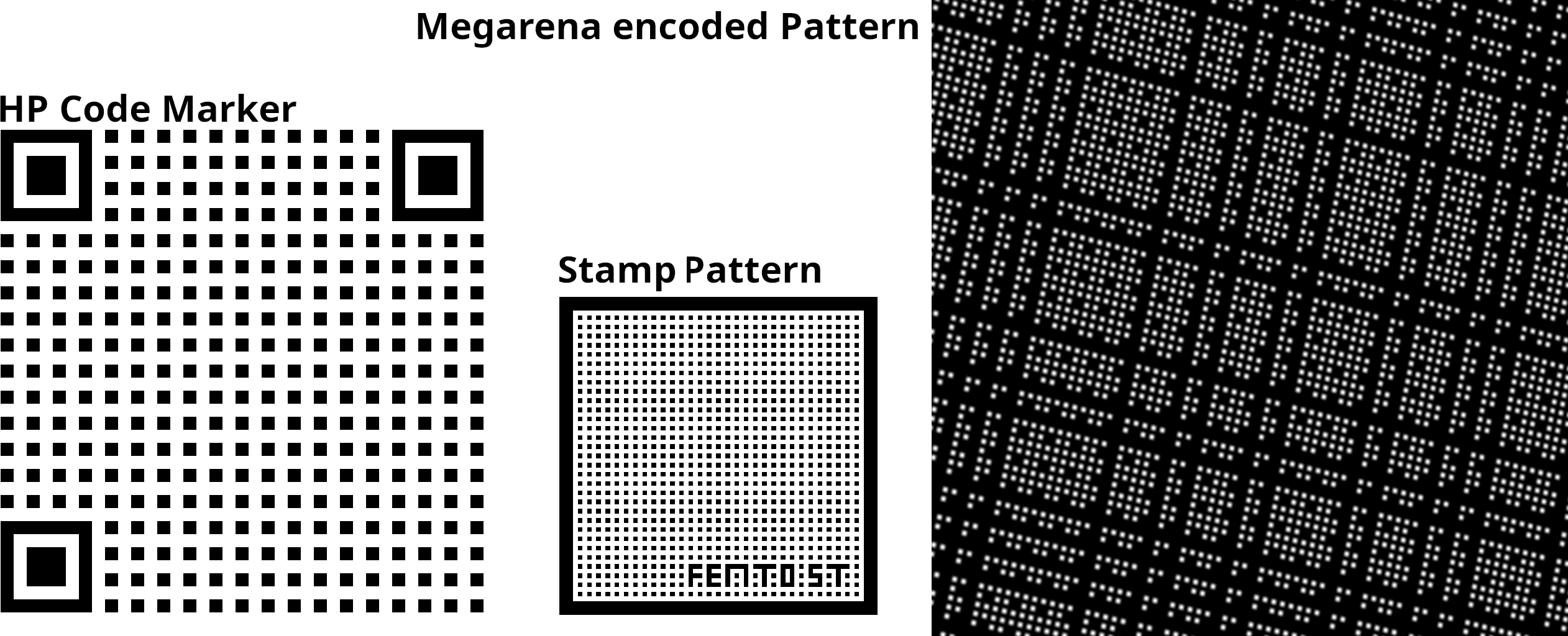}
    \caption{The three marker types (on the left an HP code marker, in the middle a Stamp marker, on the right a Megarena encoded pattern).}
    \label{fig:MarkerTypes}
\end{figure}

The small marker family consists of two designs, which differ in the method of preliminary coarse position detection. Firstly, HP code markers, where HP stands for High Precision (Fig.~\ref{fig:MarkerTypes} on the left), are based on the QR code design and reproduces the same corner squares to be suitable for QR code detection methods. The second design is called Stamp marker (see Fig.~\ref{fig:MarkerTypes} in the middle), which is detected through a polygon detection method. Both designs are filled with periodically spaced dots and overall designs are not symmetrical in order to retrieve the absolute orientation of the markers. Missing dots can be used to distinguish between different markers in a single image by encoding a marker number.

Large markers are named Megarena and are designed as an altered 2D periodic set of dots (Fig.~\ref{fig:MarkerTypes} on the right). The periodic frame aims to provide a sharp spatial frequency allowing band-pass filtering and noise rejection to achieve high resolution. The missing dots altering the periodic frame fulfill two complementary functions. One the one hand, one missing dot for each set of $3\times3$ avoids any $\pi/2$ rotation ambiguity, thus allowing absolute orientation measurement. On the other hand, full lines and columns of missing dots are used to encrypt pseudo-random binary sequences and to encode the position of the observed zone within the entire Megarena area. The encoding method is based on linear shift register sequences and further details can be found in~\cite{andre2020sensing}.

\subsection{Measurement principle}
\begin{figure}
    \centering
    \includegraphics[width = \linewidth]{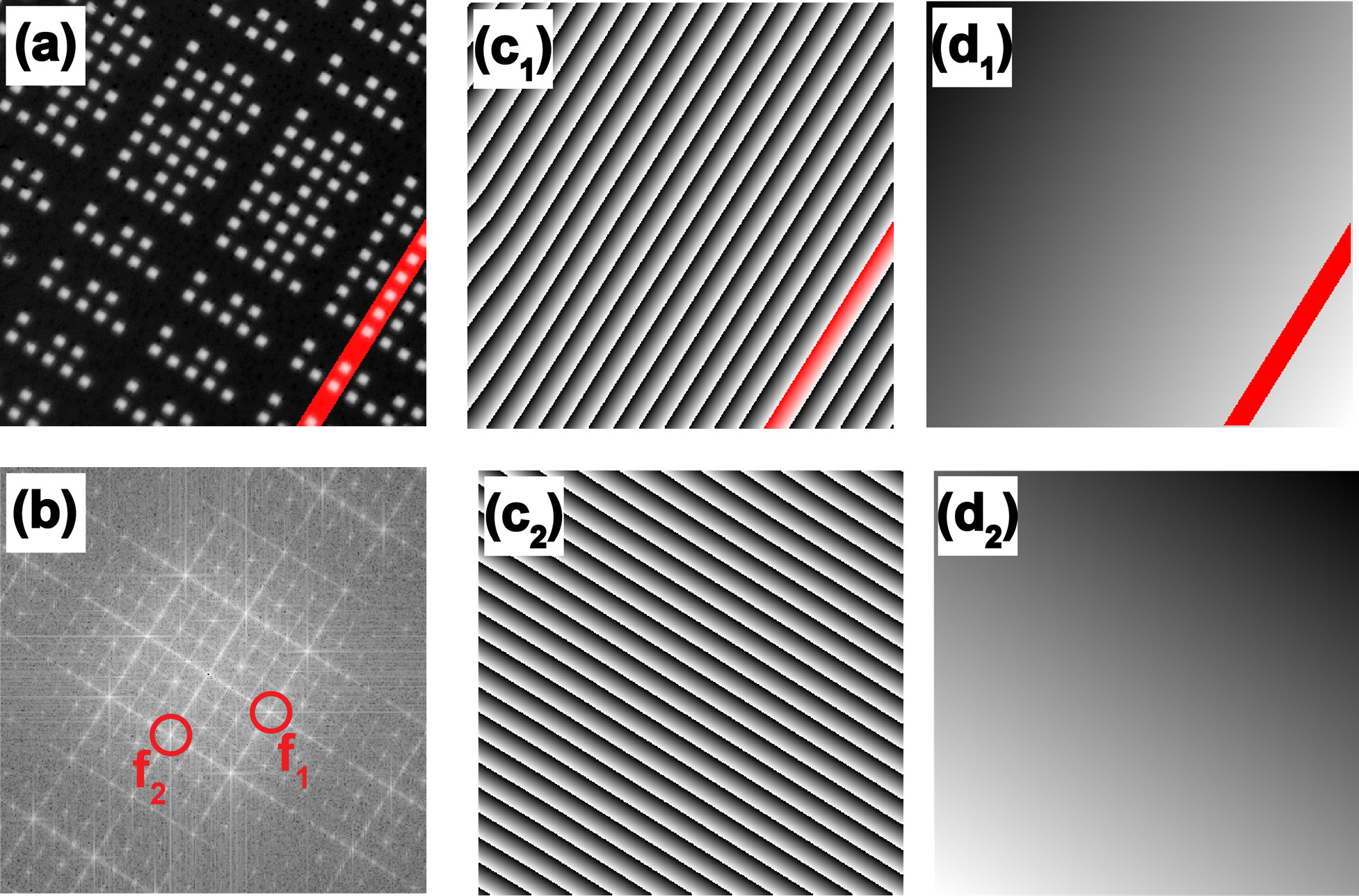}
    \caption{Fine localization process of Megarena patterns: a) recorded image; b) Fourier spectrum; c$_{1,2}$) 1D wrapped phase maps obtained from spectral lobes $f_1$ and $f_2$, respectively, through inverse Fourier transform; d$_{1,2}$) unwrapped phase planes encode the fine position of the periodic frame of dots with respect to the image pixel frame. The red stripe identifies a single line of dots for the sake of easier understanding.}
    \label{fig.principle}
\end{figure}

The pose estimation combines two complementary steps; i.e. a coarse but absolute one and a fine but relative one. The fine measurement principle ensures the high resolution of the method that is typically of a few $10^{-3}$ image pixel. It relies on Fourier transform and phased-based measurements after spectral frequency filtering. The successive processing steps are depicted in Fig.~\ref{fig.principle}.
The acquired image (Fig.~\ref{fig.principle}.a) is first transferred to spectral domain through Fourier transform and the periodic frame of dots will result in sharp spectral peaks (Fig.~\ref{fig.principle}.b). A Gaussian band-pass filter is then applied to each spectral lobe $f_1$ and $f_2$ to retrieve their frequency and phase information independently without loss of information. 
Applying an inverse Fourier transform to the two selected spectral lobes yields two wrapped phase maps (Fig.~\ref{fig.principle}.$c_{1,2}$), which can be unwrapped to obtain two phase planes that are representative of the dot positions with respect to the image pixel frame. Data averaging over the whole detected area ensures noise rejection and high resolution. However, a $2k\pi$ (with $k$ an entire number) phase ambiguity remains since the unwrapping process depends on the starting pixel chosen. This ambiguity is resolved through further absolute decoding. 
Each period of dots corresponds to a phase excursion of $2\pi$ that can be easily identified from the unwrapped phase maps. The in-plane orientation of the dot pattern is also provided by the unwrapped phase map equation obtained through least square plane fitting. 

\begin{figure}
    \centering
    \includegraphics[width = \linewidth]{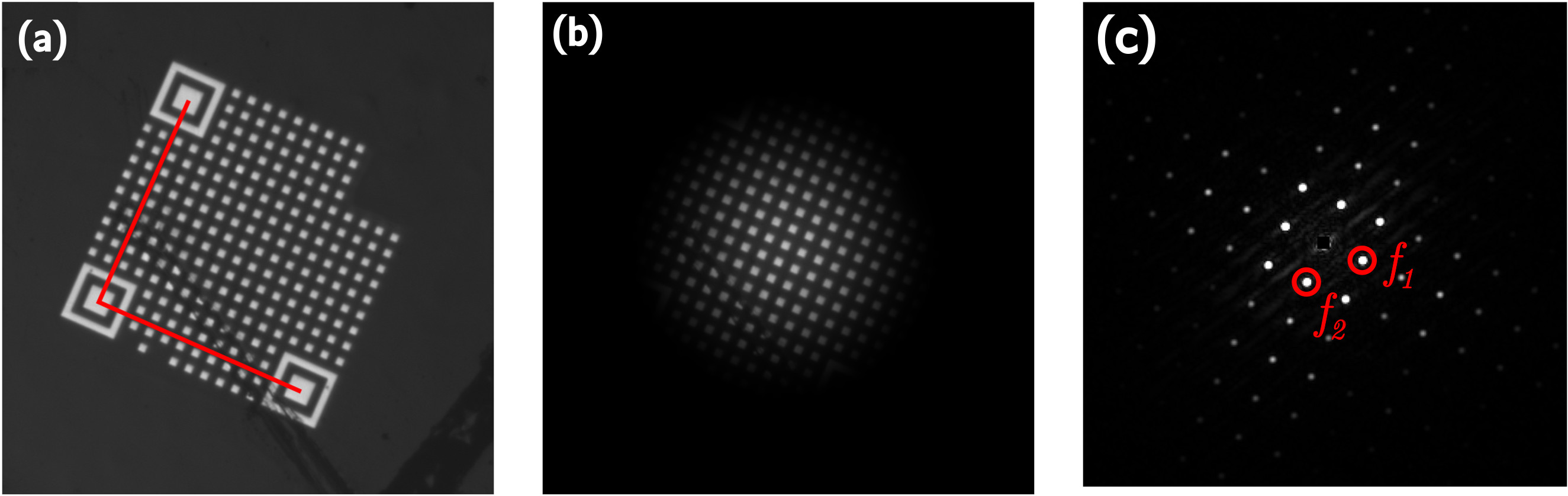}
    \caption{Localization process of HP codes: a) corners detection in recorded image; b) image apodization; c) peaks detection in Fourier spectrum.}
    \label{fig.principleHP}
\end{figure}

The coarse but absolute position detection principles differ for the small and large marker types. For small markers, we used existing robust detection methods provided by the OpenCV library; i.e. QR code detection for HP code markers (as shown in Figure~\ref{fig.principle}) and quadrilateral detection for Stamp markers. The detection of the marker contours allows the resolution of the $2k\pi$ phase ambiguity and results leading to the fine and absolute pose estimation within the image. For the Megarena marker, a robust phase-based binary decoding procedure has been developed that computes a local adaptive threshold for discriminating between present and absent dots. We thus reconstruct two binary sequences, one for each direction of the pattern, that allows the univocal determination of the position of the area observed with respect to the top-left corner of the whole Megarena pattern. 
A complete presentation of the position decoding method can be found in~\cite{andre2020robust}.

This measurement principle is mainly suited for in-plane 3 DoF pose estimation under microscopy orthographic projection. However, long-focal  perspective projection can be used for retrieving complementary out-of-plane pose parameters with a lower resolution. Full out-of-plane pose estimation details and performances can be found in~\cite{andre2022pose}. 

To achieve the measurement of the 6 DoF, we applied the same approach with a digital holographic microscope (DHM). The interferometric character of DHM makes the device highly sensitive to out-of-plane motion and the 6 DoF are measured simultaneously with a high resolution~\cite{ahmad2024-6DoF}. 

\begin{figure*}
    \centering
    \includegraphics[width = 0.95\linewidth]{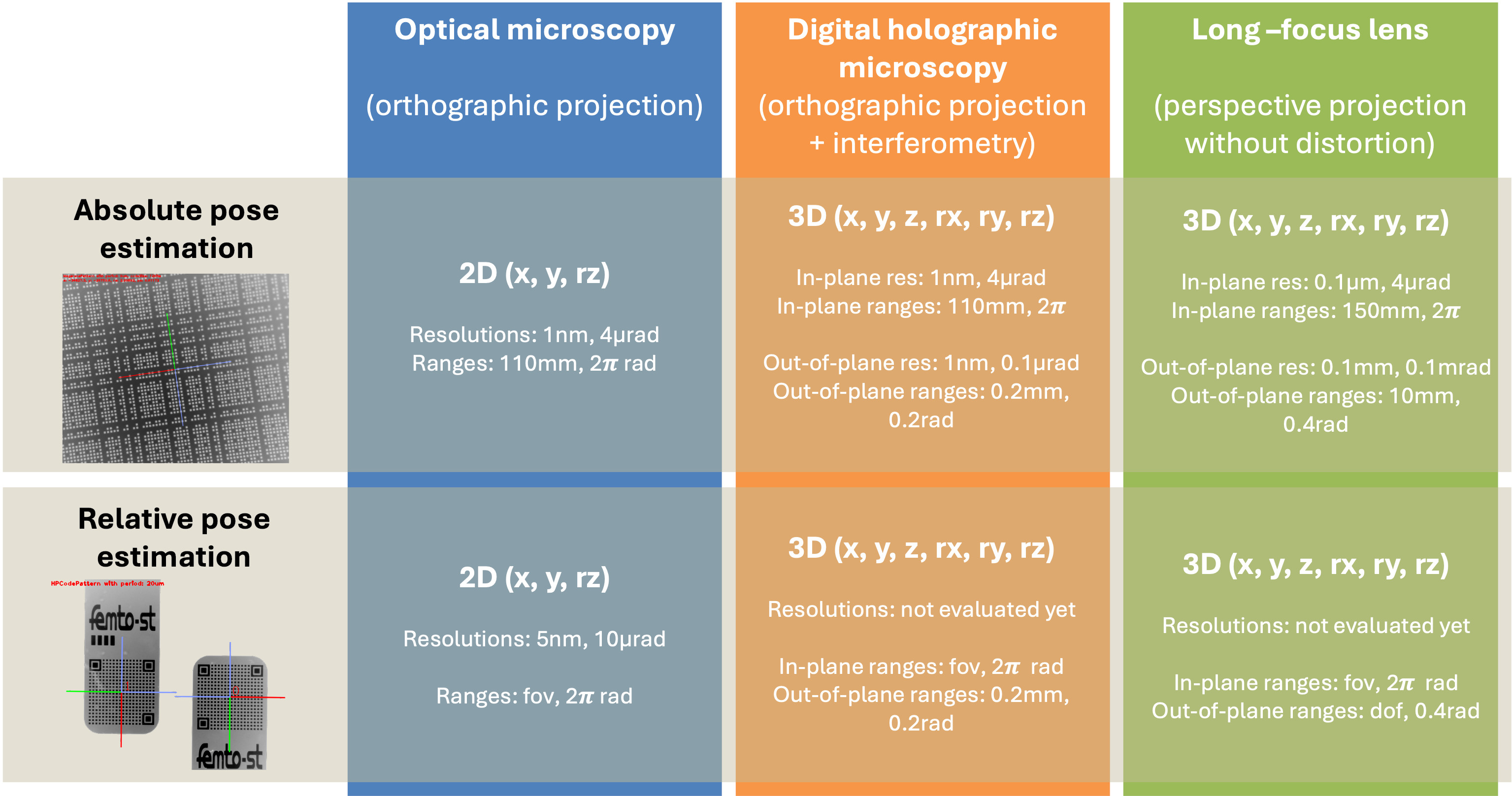}
    \caption{Overview of features and metrics of the marker pose estimation with VERNIER.}
    \label{fig:features}
\end{figure*}

Fig.~\ref{fig:features} presents the mean features and metrics of the pose estimation of both marker types. 
All the metrics have been validated experimentally with precision stages and robots.

\section{Library description}\label{sec:library}

VERNIER is written in \texttt{C++}. It defines a collection of classes for detection and rendering the different kind of calibrated patterns (periodic patterns, Megarena patterns, HP code and Stamp markers). The library is cross-platform and users can build and run the examples and unit tests on their platform/compiler. The library uses CMake as a general build tool and Doxygen to generate the documentation. The library relies on multiple third parties: OpenCV, Eigen, FFTW, MatIO, RapidJSON and GDS Tool Kit.

The library is straightforward to use and gives directly the transformation matrices between the camera frame and the marker frame.

The library also provides a set of classes for rendering synthetic images and exporting layouts of the different markers presented in previous section. The synthetic images are used to test the correct functioning of the detectors. The marker layouts can be exported in PNG, SVG, GDS and OASIS formats. These files can be used to print the markers on various supports.

As the measurement method is based on the phase of the periodic pattern, the position accuracy is directly related to the scale and quality of the periodic dot frame. However, the method is only dependent on the physical period between the dots and the size of the dots themselves does not affect the measurement. 
To obtain calibrated measures, all our patterns have been realized using a high-resolution maskless aligner (Heidelberg MLA150) on quartz or glass substrates within the MIMENTO facility\footnote{MIMENTO is a node of the French RENATECH network of large-scale facilities dedicated to technological research in micro and nanotechnology. 
See \url{https://platforms.femto-st.fr/centrale-technologie-mimento/} for more information.}. 

\section{Applications}\label{sec:applications}

\begin{figure*}
    \centering
    \includegraphics[width = 0.95\linewidth]{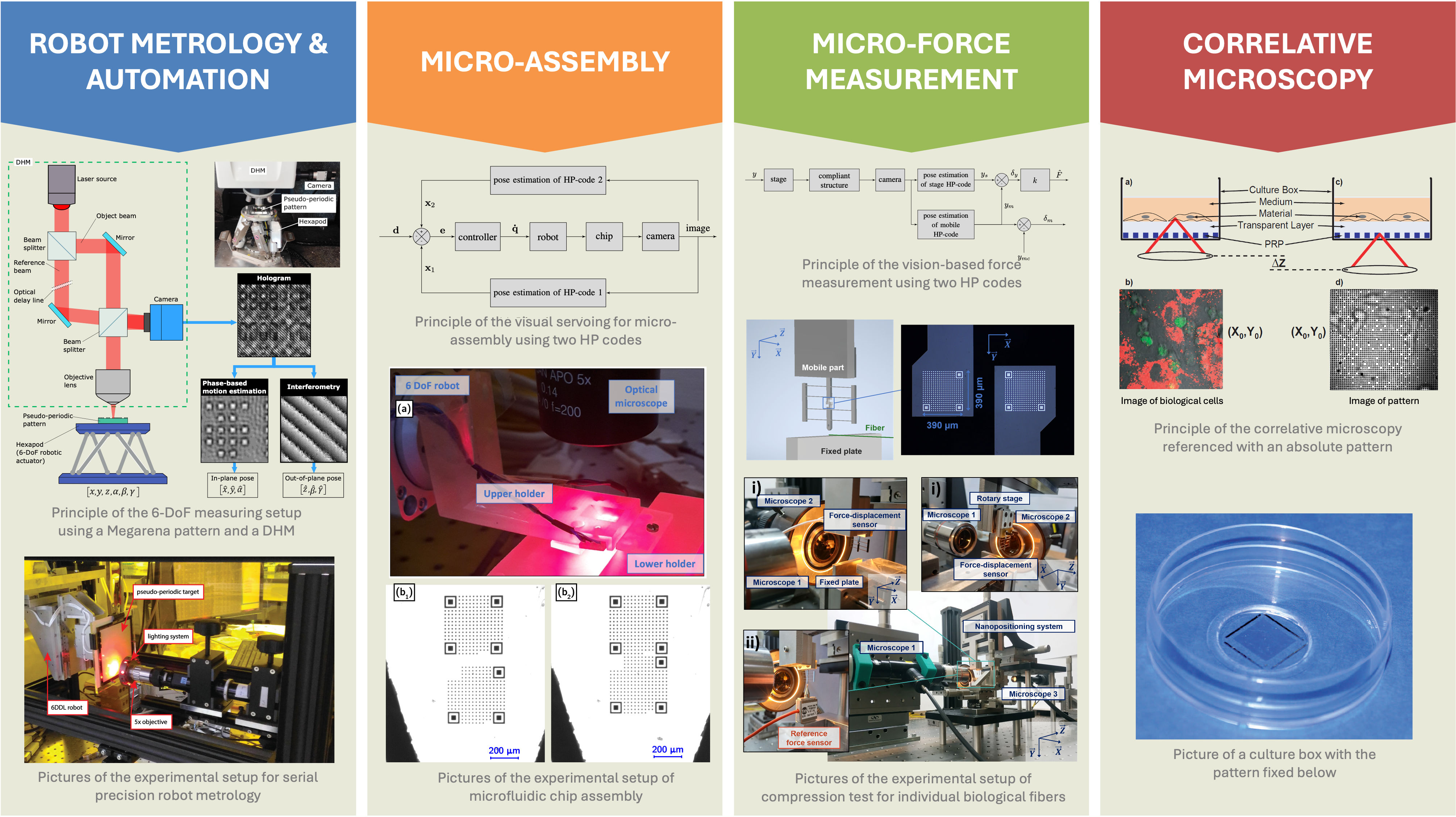}
    \caption{Overview of applications of pose estimation at the small scales with VERNIER.}
    \label{fig:applications}
\end{figure*}

In this section, we review some applications of VERNIER within the FEMTO-ST Institute and other potential applications in different fields. An overview of these applications are presented in Fig.~\ref{fig:applications}.

\subsection{Metrology of precision manipulators}

One of the major interest of Megarena patterns is to perform the metrology of micro and nano stages and precision manipulators. Indeed, few solutions to measure the 3D pose of the manipulator end-effector at the nanoscale are available. The laser interferometers provide a very high range-to-resolution ratio of approximately $10^{9}$ but only along the laser's axis. Setups with several interferometers have demonstrated multiple DoF measurement systems \cite{lee2011design, ortlepp2024high}, at the expense of occupied volume and calibration complexity. 
Moreover, due to the constraints of laser reflection, the range of their angular measurements is very low, not exceeding a magnitude of one milliradian.

The use of Megarena patterns is much simpler and also provides nanometric resolution and centimeter ranges. Moreover, using a DHM, the angular ranges of out-of-plane rotations reach 0.2\,rad with resolutions down to 0.1\,\textmu{}rad. 
The ideal setup consists only of a microscope and a piece of a Megarena pattern attached to the end effector of the manipulator. 

For example, the method has been used to evaluate the accuracy of a precision hexapod, showing that it is able to follow millimeter trajectories with deviations of less than one micrometer \cite{ahmad2024-6DoF}. It has also been employed to evaluate the accuracy of a serial 6-DoF manipulator from SMARACT over a point matrix of 10 by 10 centimeters \cite{andre2020sensing}. 
Megarena patterns can also be used to measure the repeatability of positioning of stages and manipulators. For instance, in \cite{Gallardo2021turning, Mauze2020nanometer}, the method is used to evaluate the precision of parallel continuum robots.

Beyond stages and manipulators metrology, the method could also be useful for measuring the deviation of CNC machines and vision measuring machines.

\subsection{Multi-DoF stage automation}

The Megarena patterns can also be used as position sensor to directly control the position of the motion platform of a multi-DoF manipulator during its operation. This ensures the actual position of the end-effector regardless of the errors introduced by assembly of stages, guidance, compliance, and backlash in the mechanical axis. It could also help to identify and correct cross-axis coupling and unwanted motion in multi-DoF systems, as proposed in \cite{Tan2015accuracy}. 

As no internal additional sensors are required, pattern-based direct measurement can be applied to an existing manipulator to improve its accuracy and to carry precision tasks such as micro-assembly. 

\subsection{Micro-assembly}

Micro-assembly requires two parts to be positioned relative to each other with high precision. To make assemblies with errors lower than a micrometer, the joint sensors of the manipulators are not sufficient and it is necessary to use visual servoing to control the relative position of the parts. To that extent, markers are placed on each of the parts to assemble and their relative position is monitored in real time during the assembly process. 
In \cite{andre2022automating}, we used two HP code markers to align and bond two parts of a microfluidic chip. This automated assembly achieved a positioning accuracy of less than 50\,nm.

Beyond this assembly example, HP code and Stamp markers could also be useful for wafer alignment in photolithography processes, die assembly in the semiconductor industry, and stitching in high-resolution 3D printing and laser processing.

\subsection{Micro-force measurement}

Another key application in microrobotics is simultaneous force and displacement sensing. Force sensing has always been an issue for micro- and nanoscale applications. Force ranges from a few mN to a few hundred mN are typically required for manipulation, assembly and characterization tasks in most fields.

In \cite{andre2022automating}, two HP code markers have been attached to a compliant mechanism. The relative motion between the markers gives an image of the compression force, while the relative position of the markers to the camera frame gives the displacement of the mechanism base. This force sensor has been used experimentally to perform automated compression tests on individual synthetic and natural fibers, providing very rich information and allowing to better understand the mechanical fiber behavior \cite{govilas2024mechanical}.

Previous experiments with this method has shown resolution measurements down to 50\,nN with experimental ranges of 50\,mN \cite{Guelpa2015vision}. The same principle has also been applied to multiple directions force sensing in \cite{Tiwari2021high}.

\subsection{Correlative Microscopy}

The pattern-based position measurement was initially developed to perform the repositioning of Petri dishes under a microscope for live cell monitoring \cite{galeano2011position}. Thanks to a pattern similar in principle to the Megarena, regions of interest of a culture dish was easily retrieved after transfers from a cell incubator to the microscope stage. Thanks to this approach, images of single cells can be captured at different time steps in a repeatable manner, as the cells can be identified based on their absolute position in the Petri dish.

In this way, Megarena patterns could also be used in correlative microscopy, to retrieve the position of the very same cell or tissue area and image it with various modalities (e.g. electron, confocal, fluorescent microscopy).

\section{Guidelines}

Since the pose estimation method is based on the analysis of the pattern image spectrum, the number of periods appearing on the image affects the sharpness of the spectral lobes, and hence the resolution of the measure. In \cite{andre2022automating}, we studied the effect of the number of HP code periods over the resolution. It appears in ideal conditions that viewing a minimum of 17 periods is required to achieve 0.001\,pixel resolution for translations and 0.1\,mrad for rotations (around $Z$ axis). However in real conditions, a larger number of periods will provide a greater robustness against occlusion and other disturbances of the image. By experience, 20 to 30 periods is a good choice for HP code and Stamp markers. 
In the case of Megarena patterns, the minimum number of periods to have in the field of view is determined by the depth of the absolute position code. For a $n$ bits pattern, $3(n+1)$ visible periods are required at minima.

The second requirement to achieve best results is that the apparent size of a period in the image should be between 7 and 15 pixels. Therefore, given the magnification of the lens and the pixel pitch of the sensor, it is easy to retrieve the physical size of the period to obtain this apparent size. Conversely, it is possible to choose an optical setup knowing the physical size of the period. 
If robustness against blur is important for the targeted application, a larger period is more suitable since a lower spectral frequency will be cut at an increased defocus distance. Furthermore, larger periods allow the choice of lower lens magnifications, with lower numerical apertures and increased depths of focus, making the measure more robust against defocus.

Once the magnification and pattern design have been set, we showed in \cite{andre2020robust} that the most influential parameter on the resolution is the dynamic range of the image. Best results are obtained with 12-bits sensors with a high contrast between the black background and the white dots. However, image saturation must be avoided to keep the imaging of dot edges as linear as possible. Moreover in case of moving targets, a sensor with global shutter is essential to avoid any motion bias.

Finally, nanoscale measurements require additional precautions to avoid drifts and vibrations. To reach a low level of mechanical disturbances, the setup must be mounted onto a heavy stand placed on an anti-vibration table like any coordinate measuring machine. It is preferable to place the microscope tube flat on the stand.   
Other uncertainties, such as thermal drift, can be mitigated by placing the system in a metrology room where the temperature and humidity are controlled. Then, the main remaining source of uncertainty comes from the heating of the camera. A long warm-up of the camera must be performed before measurements. These guidelines are summed up in \cite{Mauze2020visual}.

\section{Conclusion}

VERNIER covers a wide range of applications requiring pose estimation at small scales and is off-the-shelf available to be easily used on these applications. Megarena patterns provide measurements with high range-to-resolution ratios for stages metrology, manipulators automation and correlative microscopy. On the other hand, HP code and Stamp markers can be used to obtain the relative position of multiple fiducial markers in the same image for assembly and force measurement.  
Currently, a 1024x0124 pixel image takes 150ms to be processed on an 2.3 GHz Intel Core i9. Porting the library to GPUs would enable the image acquisition framerate to be reached, allowing VERNIER to be used for many automation tasks. 

\bibliographystyle{IEEEtran}
\bibliography{paper}

\end{document}